\documentclass[letterpaper, 10pt, conference]{IEEEtran}
\usepackage{times}
\usepackage{xr}
\usepackage{amsmath,amssymb,mathrsfs}
\usepackage[]{algorithm2e}
\usepackage{graphicx}
\usepackage{xcolor}
\usepackage{wrapfig}
\usepackage{textcomp}

\definecolor{control}{RGB}{203, 65, 107}
\definecolor{shape}{RGB}{61, 153, 115}

\usepackage[numbers]{natbib}
\usepackage{multicol}
\usepackage[bookmarks=true, hidelinks]{hyperref}

\begin{document}

\title{Automated shapeshifting for function recovery in damaged robots}

\author{Author Names Omitted for Anonymous Review. Paper-ID [add your ID here]}


\author{

\IEEEauthorblockN{Sam Kriegman$^1$,  Stephanie Walker$^2$,  Dylan Shah$^2$, Michael Levin$^3$,  Rebecca Kramer-Bottiglio$^2$, Josh Bongard$^1$}
\IEEEauthorblockA{$^1$University of Vermont,  $^2$Yale University,  $^3$Tufts University}


}

\teaser{
\centering
\vspace{-6pt}
\includegraphics[width=\linewidth]{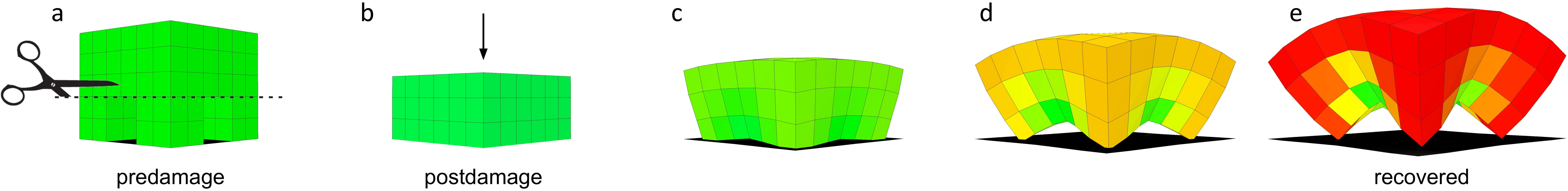} \\
\includegraphics[trim={0 0 0 4pt},clip,width=\linewidth]{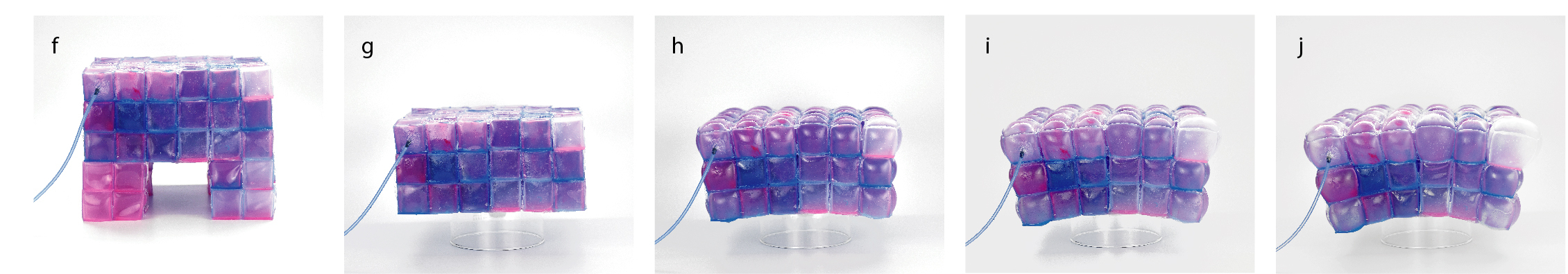} \\
\vspace{-4pt}
\caption{After learning to walk, a simulated quadruped is subjected to unanticipated insult: its legs are cut off. 
An evolutionary algorithm searches for deformations to the postdamage structure that, when coupled with the predamage controller, result in function recovery.
One of the evolved solutions (shown here) yields the spontaneous ``regeneration'' of the lost legs, which was manually transferred to reality (\href{https://youtu.be/afOXX2r54mQ}{\textcolor{blue}{\textbf{\texttt{youtu.be/afOXX2r54mQ}}}}).
} 
\label{fig:teaser}
\vspace{-20pt}
}

\maketitle

\begin{abstract}
A robot's mechanical parts routinely wear out from normal functioning and can be lost to injury. For autonomous robots operating in isolated or hostile environments, repair from a human operator is often not possible.
Thus, much work has sought to automate damage recovery in robots.
However, every case reported in the literature to date has accepted the damaged mechanical structure as fixed, and focused on learning new ways to control it.
Here we show for the first time a robot that automatically recovers from unexpected damage by deforming its resting mechanical structure without changing its control policy.
We found that, especially in the case of ``deep insult'', such as removal of all four of the robot's legs, the damaged machine evolves shape changes that not only recover the original level of function (locomotion) as before, but can in fact surpass the original level of performance (speed).
This suggests that shape change, instead of control readaptation, may be a better method to recover function after damage in some cases.
\end{abstract}

\IEEEpeerreviewmaketitle

\section{Introduction}
\label{sec:intro}

Certain remote, hazardous or otherwise inaccessible environments preclude human intervention when a robot fails or is damaged.
It would thus be advantageous for systems operating in such environments to have some capacity for self -maintenance and -repair.

Indeed, much work has investigated how,
in the absence of external supervision,
a robot can automatically learn new ways to control its body when damaged~\mbox{\cite{bongard2006resilient, chatzilygeroudis2018reset, cully2015robots, kano2017brittle, kwiatkowski2019task, mahdavi2003evolutionary, ren2015multiple}.}
While a diverse set of recovery mechanisms have been proposed, they all  
shared a common assumption: 
The damaged mechanical structure could be 
reconfigured, but not fundamentally deformed.

This assumption is reasonable in classical robots, which are, generally, jointed collections of rigid links.
But recent advances in materials science and 3D printing are enabling the construction of soft machines with theoretically infinite degrees of freedom and thus capable of deforming their
structures so as to regenerate a lost part or embrace an entirely new geometry in the face of unanticipated insult. 
The possibility of such change affords a completely novel mode of damage recovery:
No robot built to date has altered its resting structure in order to recover function lost due to damage.

Previous computational studies have demonstrated structural but non-functional change in discrete models.
For example, cellular automata have been trained to grow a target structure from a single cell \cite{eggenberger1997evolving, miller2004evolving}.
Similar growth rules could in principle be instantiated in self-assembling modular robots~\cite{white2005three,zykov2005robotics}.
However, structural change would require access to additional modules in the environment, redundant modules on the body, or the ability to internally generate them.
Moreover, it is unclear how or if such rules could dictate continuous geometric deformation in soft robots.

The present work builds on two closely-related research projects in which injured robots automatically generate and test candidate control policies in order to find compensatory behaviors that work in spite of damage
\cite{bongard2006resilient,cully2015robots}.

In the first, \citet{bongard2006resilient} demonstrated how, under the right conditions, an autonomous robot could internally model its own geometry with minimal sensorimotor experiment.
The benefit of this approach is that, once a sufficiently accurate self-model has been established, actions can be internally rehearsed, discarding those which are unsuccessful or dangerous, before attempting them in reality.
If model accuracy drops, as from structural changes due to damage, modeling resumes and continues until the robot's current morphology is adequately reflected in the robot's model of self.

The main drawback of this approach is that internal modeling requires additional computation, and there are circumstances in which the robot cannot afford|in terms of time, money, energy, stability, and the overall well-being of itself and others|to remain stationary for extended periods of time.

To speed recovery, \citet{cully2015robots} proposed that robots should instead exploit the fact that resources prior to deployment are relatively cheap in terms of the factors listed above.
A large, behavioral repertoire composed of mappings from behaviors (for the undamaged robot) to their predicted performances can therefore be modeled in simulation beforehand, and come preinstalled on the robot.
Assuming damage is detected by an external mechanism, the authors showed how, under certain conditions, such a map can be rapidly updated and traversed to find successful behavior, 
which is 
implicitly robust to differences between the current and pre-deployment
morphologies.

The robots used in this past work consisted of rigid components attached together with a handful of mechanical degrees of freedom:
The quadruped in \cite{bongard2006resilient} had 8 motors and 4 DOF; 
the hexapod in \cite{cully2015robots} had 18 motors and 12 DOF.
The control problem was greatly impacted by these mechanical details and their intrinsic dynamics, but they were taken as given, even when damaged, because these robots simply could not deform their resting structure.

Instead of treating the body as just the problem domain, we here modify it as part of the computational loop.
This is possible because our robot has many more (140) mechanical degrees of freedom, and the ability to change the volume, rather than just the relative displacement, of each component.
This flexibility enables a heretofore unexplored mode of damage recovery: keep the existing controller but deform the resting structure.
Existing approaches to controller adaptation could in principle (although this is not investigated here) be paired with such changes to morphology.
However, in many cases, it would be desirable to retain a previously optimized and fine-tuned controller, especially if missing structure can simply be regenerated.

We here show that, under a wide range of damage scenarios, automated shapeshifting can be advantageous, and that, in most of the cases tested, shapeshifting alone (holding the existing controller fixed) outperforms controller adaptation alone (holding the damaged shape fixed), in terms of recovered mobility.

\section{Methods}
\label{sec:methods}

This section describes the hardware, simulation and control of our robot,
the damage scenarios it faces and its options for recovery: 
shapeshifting and controller adaptation.
We also define a tripartite classification|of `structure', `shape' and `configuration'|that forms the basis of our argument, which is, briefly, that the way in which our robot recovers from damage|shape change|was outside the scope of any robot previously reported in the literature.

\subsection{The source code.}
\href{https://github.com/skriegman/2019-RSS}{\textcolor{blue}{\textbf{\texttt{github.com/skriegman/2019-RSS}}}}

\subsection{The robot.}
\label{sec:robot}

\begin{wrapfigure}{r}{0.5\linewidth}
\vspace{-1em}
\includegraphics[trim={0 16em 0 17em},clip,width=\linewidth]{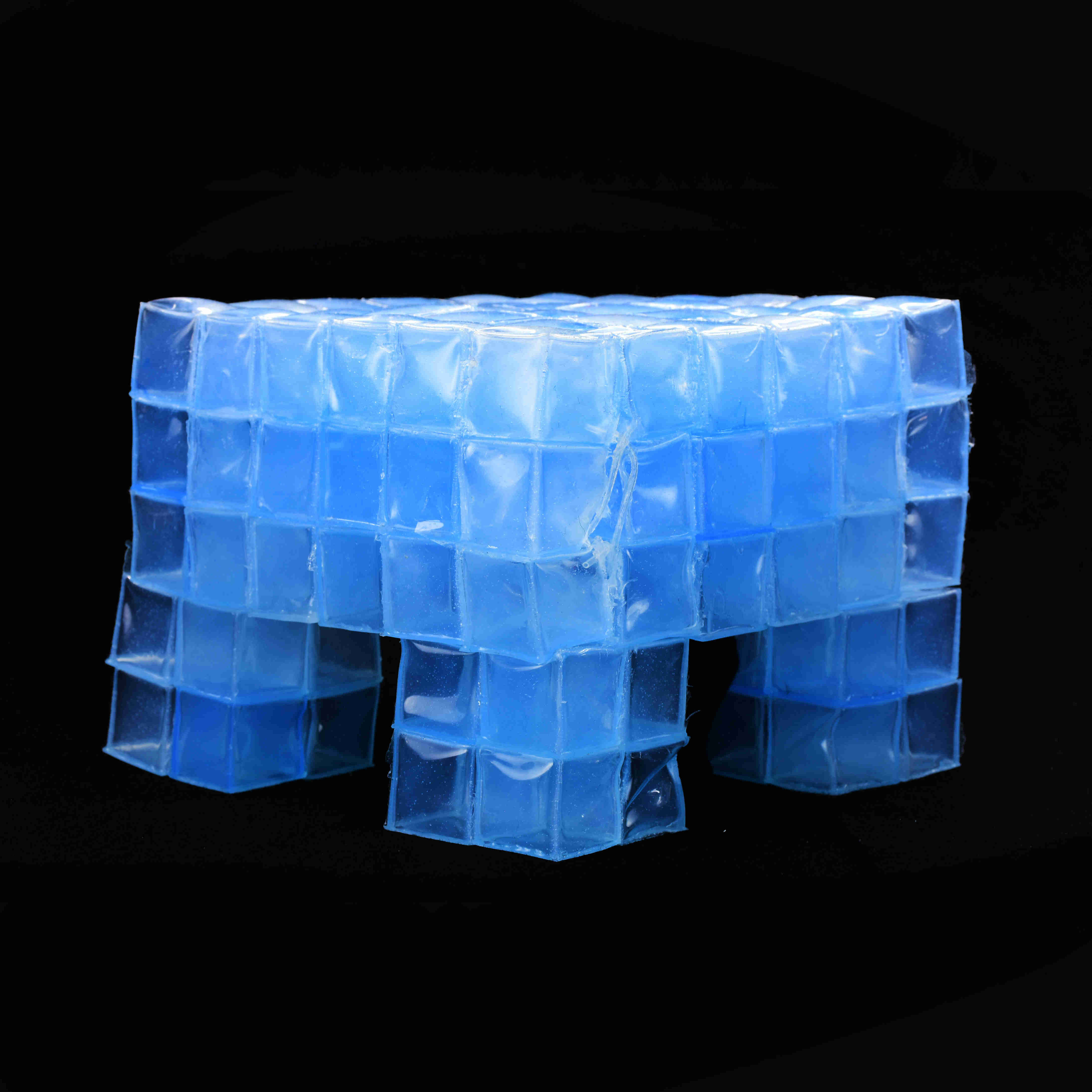}
\vspace{-18pt}
\caption{The blue robot, made from thin-walled inflatable elastomer voxels.}
\label{fig:blue_quad}
\vspace{-1em}
\end{wrapfigure}

The robot is an isobilaterally symmetrical quadruped constructed from 140 inflatable silicone ``voxels'' 
(Figs.~\ref{fig:teaser}f and~\ref{fig:blue_quad}).
We here present a method for creating air-filled voxel membranes with relatively uniform thickness.

Creating thin, hollow 3D silicone structures is challenging due to several factors, including mold precision and potential for damage during release from molds. One effective but labor-intensive method is to make the 3D shapes by adhering 2D films at their joints \cite{morin_elastomeric_2014}. Here, inspired by a scalable 2-axis rotational molding technique \cite{zhao_scalable_2015}, we employ a \mbox{1-axis} rotational drip-molding machine.

First, silicone (Dragon Skin 10 Fast; Smooth-On, Inc.) was poured into an open-face acrylic mold and a tongue depressor was used to roughly spread the silicone along the walls. The mold was then attached to the rotational molding machine with the rotation axis oriented downward at 45 degrees relative to horizontal, and run through cycles comprising a 90\textdegree~turn, stopping for 45 seconds after each turn to allow the silicone to flow and evenly coat each side. Excess silicone dripped out of the mold, leaving a thickness which was dependent on several interrelated factors including the cure time, viscosity, and the interaction between the silicone and acrylic.

After the silicone cured, excess material was cut away. A silicone base-layer was then rod-coated onto a flat acrylic sheet. Next, the bottomless cubes were placed on the base-layer and allowed to cure, sealing air inside each voxel. The voxels were then cut from the sheet and a small hole was punched in each voxel for tubing. Finally, silicone tubes were inserted and bonded with Sil-Poxy (Smooth-On, Inc.).

The overall robot consists of a $6\times6\times3$ voxel torso and four removable $2\times2\times2$ voxel legs (Figs.~\ref{fig:teaser}f-j  and~\ref{fig:blue_quad}). 
Sil-Poxy and Ecoflex 00-50 were used to improve adhesion between voxels. 
To explore the effect of layer thickness on the range of attainable morphologies,
two versions of the robot were fabricated:
The {\color{blue}\textbf{blue robot}} (Fig.~\ref{fig:blue_quad}) consists of voxels made with one layer of silicone, while the {\color{purple}\textbf{purple robot}} \mbox{(Fig.~\ref{fig:teaser}f-j)} consists of thicker-walled voxels made with two layers of silicone. 

Individual cubic voxels were manually inflated at pressures less than 20~kPa, and approached a spherical shape as pressure increased. 
When patterned together into a robot, selective inflation of a subset of voxels induces overall robot shape change. 
To reduce friction and weight effects in the robots, they were placed on top of a glass crystallizing dish, which lifted their legs off the table surface.
While this arrangement made motion difficult, it allowed us to conduct a preliminary investigation of the feasibility of transferring simulated shape change to a physical system. 
In future implementations, the manual inflation could be replaced by pressure regulators~\cite{booth_addressable_2018}, allowing the robot to approach the continuous control achievable in simulation.

To understand some of the trade-offs between design parameters, consider a spherical pressure vessel in uniform free expansion:
\begin{equation}
    \label{eq:pressure vessel}
    p=\frac{2 E \cdot \epsilon \cdot t}{r} = \frac{2 E \cdot \epsilon \cdot t_0 \cdot (1-\delta)}{r_0-\epsilon},
\end{equation}
where $t_0$ [m] is the thickness of the pressure vessel, $r_0$~[m] is the radius, $\epsilon$ is the linear strain due to expansion, $E$ [MPa] is Young's modulus, and $\delta$ is the radial strain (which is determined from $\epsilon$ and the material's Poisson's ratio).
Note that each voxel can push outward with a force proportional to the pressure. Examining Eq.~\ref{eq:pressure vessel}, we see that at a given strain rate and initial dimensions, the internal pressure scales linearly with both thickness and modulus. Thus, when choosing thickness of voxels, there was a tradeoff between weight and internal pressure: doubling the wall thickness doubled weight, in exchange for doubled operational pressure.

\subsection{The simulation.}

\begin{wrapfigure}{r}{0.5\linewidth}
\vspace{-1em}
\includegraphics[trim={0 0 0 0},clip,width=\linewidth]{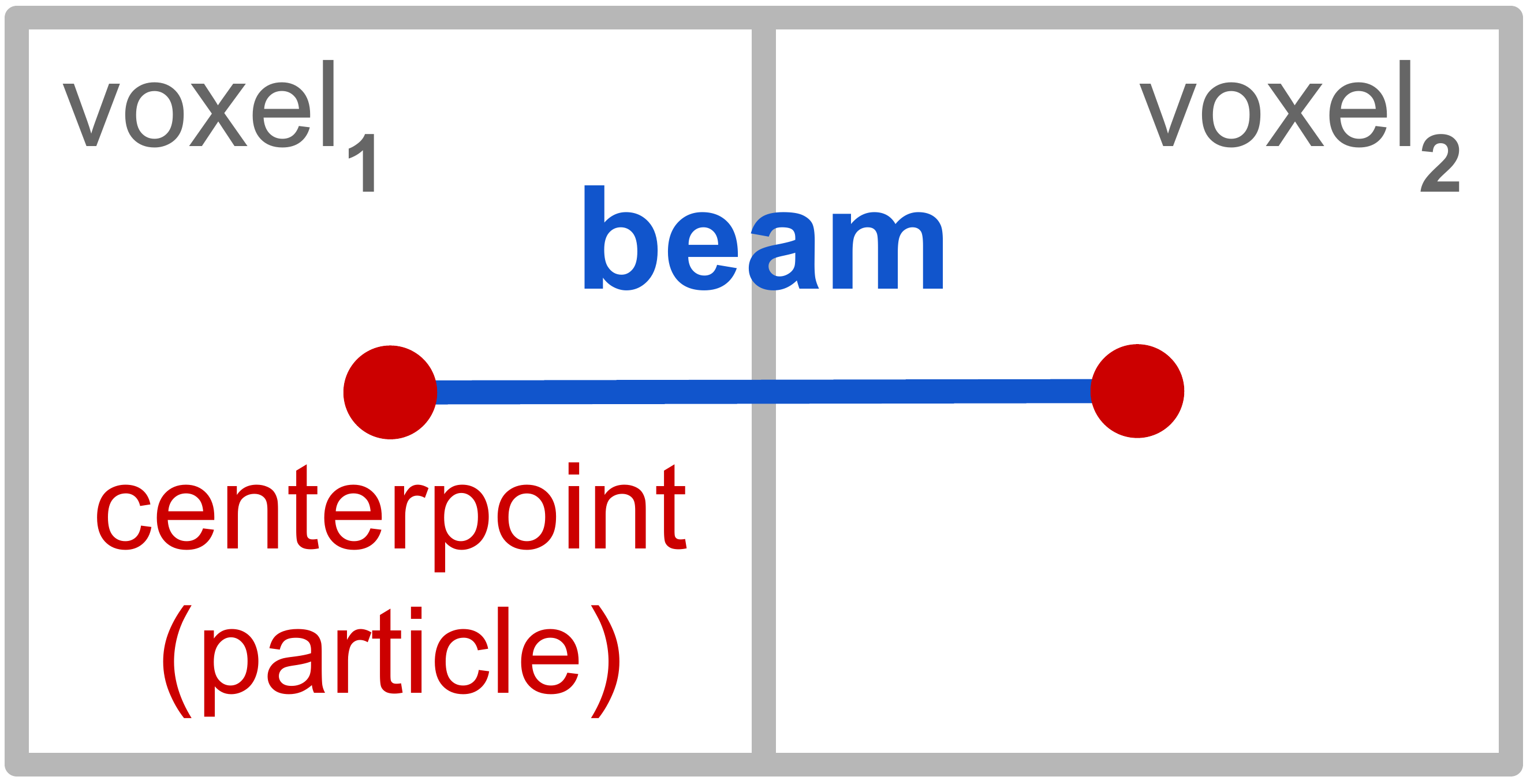}\\
\vspace{-19pt}
\caption{Voxels are simulated by beams (springs) and particles (masses).
}
\label{fig:voxcad}
\vspace{-1em}
\end{wrapfigure}

To simulate the robot, we use the voxel-based physics engine \textit{Voxelyze} \cite{hiller2014dynamic},
which simulates elastic voxels using two elements: particles and beams.
Particles have mass and rotational inertia, and are connected on a cartesian grid by spring-like beams (with translational and rotational stiffness).
For visualization and reference, part of a voxel mesh is drawn around this structure such that each voxel has a single particle at its center (Fig.~\ref{fig:voxcad}).

Two adjacent voxels are connected, centerpoint to centerpoint (i.e.,~particle to particle), by a single, shared beam.
Material properties (e.g.,~volume and elasticity) are specified at the particles but implemented as attributes of beams (e.g.,~their rest length, and how easily they twist and stretch).
Where two adjacent particles disagree in their ``desired'' attributes of a shared beam, an average is taken.

A beam exits a voxel normal to, and in the center of, one of the voxel's faces.
Although the mesh is drawn such that voxel edges bend around the underlying beam-mass network (see,~e.g.,~Fig.~\ref{fig:teaser}), a spherical envelope is used for collision detection, thus approximating the spherical expansion of the physical voxels (with maximal expansion occurring at the center of each face).
For more details see \cite{hiller2014dynamic}.

\subsection{The structure and shape of a robot.}

The \textbf{structure}, $\mathbb{S}$,
of a robot is determined by the number and placement of voxels, and simulated by the presence and absence of particles on a regular grid in the workspace.
Let the bit value $v_i$ denote the presence ($v_i=1$) or absence ($v_i=0$) of a voxel at index~$i$.
The structure,
\begin{equation}
    \label{eq:structure}
    \mathbb{S} = \{i : v_i = 1 \} ,
\end{equation}
is thus a set of voxel coordinates.

The \textbf{shape}, $\mathcal{S}$,
of a robot is determined by the resting volume of each voxel, which is expressed in simulation as the resting (or, equilibrium) lengths of the beams connecting adjacent particles, and in reality as a resting pressure within each voxel (though the exact pressure, $p_i$, is not measured here).
Let the floating point value $b_i$ denote the beam rest length stored at the $i$-th simulated voxel.
The shape,
\begin{equation}
    \label{eq:shape}
    \mathcal{S}_{\text{sim}} = \{b_i : i \in \mathbb{S} \} \; \sim \; \mathcal{S}_{\text{real}} = \{p_i : i \in \mathbb{S} \} ,
\end{equation}
is thus a set of voxel resting volumes.

The robot has a quadrupedal predamage structure (Figs.~\ref{fig:teaser}a,f and~\ref{fig:blue_quad}) with atmospheric voxel resting pressure, which is approximated by nominal beam rest lengths of 1~cm.
Damage removes structure (voxels) (Fig.~\ref{fig:teaser}b).
Postdamage structural deformation|shape change|is executed by pressure changes in the remnant structure (i.e.,~mutations in $\mathcal{S}_{\text{real}}$) \mbox{(Fig.~\ref{fig:teaser}h-j)} and approximated by local adjustments in the remaining beam-mass network (i.e.,~mutations in $\mathcal{S}_{\text{sim}}$) (Fig.~\ref{fig:teaser}c-e).
The mechanical structure and its resting shape are fixed prior to behavior during the evaluation period (20 actuation cycles).

\subsection{The controller and configuration of a robot.}
\label{sec:methods:controller}

The controllers continuously reconfigure the volume of a given mechanical structure during the evaluation period.
We here consider open loop control of 
$\pm0.5$ cm$^3$ volumetric change ($\pm50\%$ from nominal),
at each voxel, with a phase offset relative to a central pattern generator, for 4~sec.

Controllers are here encoded as neural networks that map the indices of voxels in 3D space (Eq.~\ref{eq:structure}) to a phase offset value, $\phi_i$, between $-2\pi$ and $2\pi$.
We chose this particular encoding, which is commonly referred to as a Compositional Pattern-Producing Network, or CPPN \cite{stanley2007compositional},
because spatial regularities (in structure and actuation) are known to facilitate locomotion.
(For more details about this encoding, see \cite{cheney2013unshackling}.)

The instantaneous \textbf{configuration}, 
$\mathcal{C}$,
of a robot is determined by an oscillating adjustment to the volume (and thus pressure) of each voxel,
centered around its shape $\mathcal{S}$.
In simulation, rest lengths are
periodically varying \mbox{($f=$ 5~Hz)} around their baseline, $b_i\,$,
with constant amplitude
\mbox{($A\approx$~0.145~cm)}, but damped by $\xi$.
Damping prevents contracting voxels from overlapping by decreasing their oscillation amplitude as their rest length approaches a lower bound of $b_i=0.25$~cm.

The instantaneous adjustment to the rest length of the $i$-th simulated voxel, at time~$t$, is thus:
\begin{equation}
\label{eq:beam_actuation}
\psi_i(t) = A \cdot \sin(2\pi f t + \phi_i) \cdot \xi(b_i) ,
\end{equation}
where:
\begin{equation}
\label{eq:beam_damp}
\xi(b) = \min\left[ 1,\; \frac{4b - 1}{3} \right] .
\end{equation}
The configuration,
\begin{equation}
\label{eq:beam_configuration}
\mathcal{C}_{\text{sim}}(t) = \{b_i + \psi_i(t) : i \in \mathbb{S} \} ,
\end{equation}
is thus a cyclical adjustment in the rest length between adjacent simulated voxels (implemented when computing the elastic force between them) throughout a structure $\mathbb{S}$ with shape $\mathcal{S}_{\text{sim}}$.

Although simple, open loop control has the ability to produce complex behaviors,
such as symmetrical and asymmetrical gaits (from patches of voxels that oscillate in counter-phase), or propagating waves of excitation (from a sequence of voxels with increasing or decreasing phase offsets).
Indeed, it is well known that central pattern generators in the mammalian spinal cord (and elsewhere in invertebrate systems) produce the basic, rhythmic motor patterns of locomotion, such as stepping, independently of sensory input \cite{goulding2009circuits}.

\subsection{The damage scenarios.}

We here consider nine damage scenarios|we amputate: (i) half of a leg; (ii) one entire leg, (iii) two adjacent legs, (iv) two diagonal legs, (v) three legs, (vi) all four legs; (vii) one quarter of the robot's body, (viii) one half of the body, and (ix) three quarters of the body.

\begin{figure}[h]
\begin{center}
\vspace{-6pt}
\includegraphics[trim={14pt 0 14pt 0},clip,width=\linewidth]{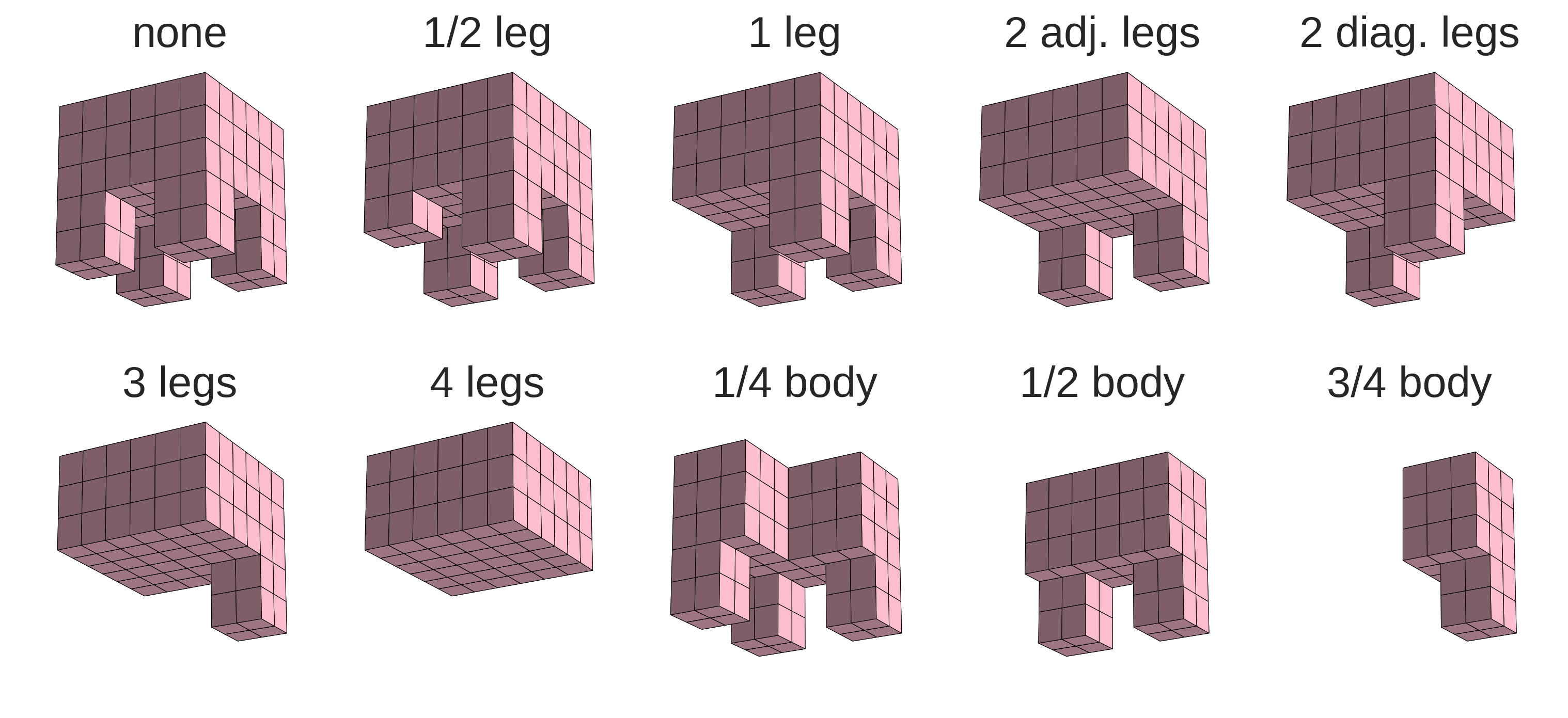}\\
\vspace{-1em}
\caption{\label{fig:scenarios}The various amputations applied in our experiments. 
The predamage robot (amputation = `none') is shown for reference.}
\vspace{-13pt}
\end{center}
\end{figure}

\subsection{The recovery options.}

Each damage scenario removes structure and breaks the robot's functionality: the robot loses voxels and its ability to walk.
We here consider two options for function recovery: 
\begin{enumerate}
    \setlength{\itemsep}{3pt}
    \item \textbf{Controller readaptation.}
    A new controller is optimized for locomotion with the damaged structure, as in \cite{bongard2006resilient,cully2015robots}.
    The only parameter subject to (re)optimization is the phase offset, $\phi_i$, of each voxel.
    \item \textbf{Shapeshifting.} The shape of the damaged structure is optimized for locomotion with the existing controller.
    The only parameter subject to optimization is the baseline rest length, $b_i$, of each voxel.
\end{enumerate}

\subsection{The shape change.}

The body is reshaped 
\textit{prior to} behavior (i.e., before the controller is turned on), analogous to a prenatal developmental stage.
This is done by adjusting the robot's shape, $\mathcal{S}$, as defined in Eq.~\ref{eq:shape}.
Then, behavior results from oscillations that are symmetrically distributed about this shape 
(Eq.~\ref{eq:beam_configuration}). 

The same kind of neural network that encodes controllers 
(i.e, a CPPN) 
was also used to encode 
the robot's shape.
However, the 
shape-encoding
networks output a rest beam length, $b_i$, between 0.25 and 2 cm (instead of a phase offset, $\phi_i$, between $-2\pi$ and $2\pi$).
Subject to the constraints outlined above,
optimization searches for shape-encoding networks that result in resting shapes that, when coupled with the original open-loop controller (previously optimized for the undamaged robot), synergize to recover forward movement.

\subsection{The optimization algorithm.}
\label{sec:optimization}

Shapes and control policies
are here optimized to displace the (simulated) robot in any direction using Age-Fitness-Pareto Optimization \citep{schmidt2011age}, 
an evolutionary algorithm that uses the concept of Pareto dominance and an objective of `age' (in addition to displacement) intended to promote diversity among candidate designs and prevent premature convergence.

A trial is initialized with a population of 50 randomly-generated designs with age zero.
Every generation, the population is first doubled by creating modified copies of each individual in the population (i.e., offspring, in which `age' is set equal to that of the parent), where modification occurs only to the encoding-network that is currently being optimized (either that of $\phi$ or $b$).
The age of each individual is then incremented by one. 
Next, an additional random individual (with age zero) is injected into the population (which now consists of 101 designs). 
Finally, selection reduces the population to its original size (50 designs) according to the two objectives of net displacement (maximized) and age (minimized): Starting with nondominated designs ($N=0$), successive Pareto fronts (containing designs dominated by exactly $N$ alternatives, for $N=1,2,\ldots$) are kept in their entirety until doing so would overfill the population past its original size; then, designs are selected one-by-one with probability proportional to their net displacement. (The 51 unselected designs are deleted.)

This process of random variation and directed selection is repeated for 
$G$ generations, in which
both the architectures and weights of the encoding networks are optimized:
Mutations add, modify or remove a particular vertex or edge.
Where modification of an edge reweights it (within -1 to 1 bounds) by adding a value randomly drawn from a normal distribution with mean zero and standard deviation 0.5.
Vertex modification swaps the node's activation function with a randomly chosen function in the set (adopted from \cite{kriegman2018interoceptive}): sin(), abs(), square(), sqrt(abs()); and the negations of those four.

\section{Results}
\label{sec:results}

Prior to damage, twenty controllers were optimized (for \mbox{$G=1500$} generations) to generate forward movement in the simulated quadruped during an evaluation period of 4 sec (with numerical integration time steps of 0.000151 sec).
Predamage displacement ranged from 37 to 46 cm (6.2 - 7.7 body lengths).

In order to isolate the effect of shape change relative to that of controller adaptation, across a diversity of insult, the simulated robot is copied $9*2*20=360$ times; once for each unique damage scenario (9 total), recovery option (2 total) and controller (20 total) triplet. 
Each copy is thus given an optimized controller and recovery option, cut according to its particular damage case, and then reoptimized for displacement (for \mbox{$G=500$} postdamage generations).

\subsection{The performance recovered after damage.}

\begin{figure}
\begin{center}
\includegraphics[width=\linewidth]{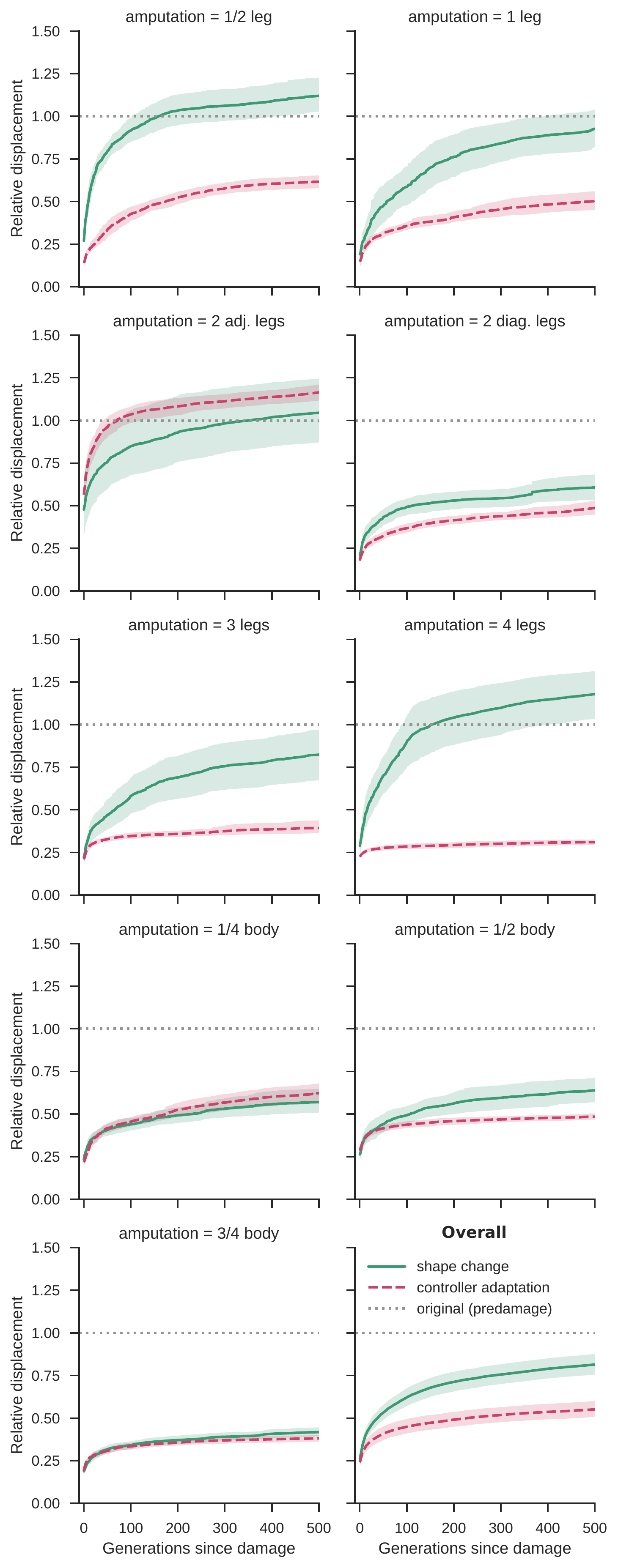}\\
\vspace{-3pt}
\caption{\label{fig:recovery}
Mean relative displacement (i.e.,~recovered performance) with 99\% CIs,
at each generation $(g)$ of reoptimization since damage occurred $(g=0)$.
}
\end{center}
\end{figure}

Figure~\ref{fig:recovery} plots 
mean relative performance (i.e.,~postdamage displacement as a fraction of predamage displacement),
with 99\% bootstrapped confidence intervals,
for the two recovery options in each damage scenario.

There are two, \textit{independent} recovery options: controller readaptation (control) and shapeshifting (treatment).
For each damage scenario,
the data consist of two random samples, a sample from the control population (20 independent trials, holding the shape of the damaged structure fixed) and an independent sample from the treatment population (20 independent shape optimization trials, holding the controller fixed).
On the basis of these samples we wish to investigate the presence of a treatment effect that results in a shift of location (median).
The null hypothesis is that of no treatment effect; 
the samples can be thought of as a single sample from one population. 

We used a distribution-free rank sum test (Wilcoxon, Mann and Whitney) for the hypothesis of no treatment effect, with Bonferroni correction for nine comparisons.
The corrected rank sum test and the 99\% bootstrapped CIs (of the mean) are in agreement.
That is, statistical significance between shapeshifting and controller adaptation, at the 0.01 level, can be correctly inferred, for each damage scenario, by visual inspection of Fig.~\ref{fig:recovery} (i.e., no overlap in the shaded confidence intervals here implies rejection of the null hypothesis).

Overall, shape change was more successful (often better and never worse) than controller adaptation.
Interestingly, the proportion of fitness recovered was in some cases higher than one.
This could be due to a lack of volume conservation and the possibility that larger robots simply run faster than small ones.
However, many robots recovered by reducing their overall volume (e.g.,~Figs.~\ref{fig:splay} and \ref{fig:scruncher}).
Moreover, controller adaptation also achieved higher-than-predamage performance in one case (amp.~=~2~adj.~legs), and this phenomenon was also documented in \cite{cully2015robots} but not via shape change.

\begin{figure}
\begin{center}
\includegraphics[trim={5pt 0 0 0},clip,width=\linewidth]{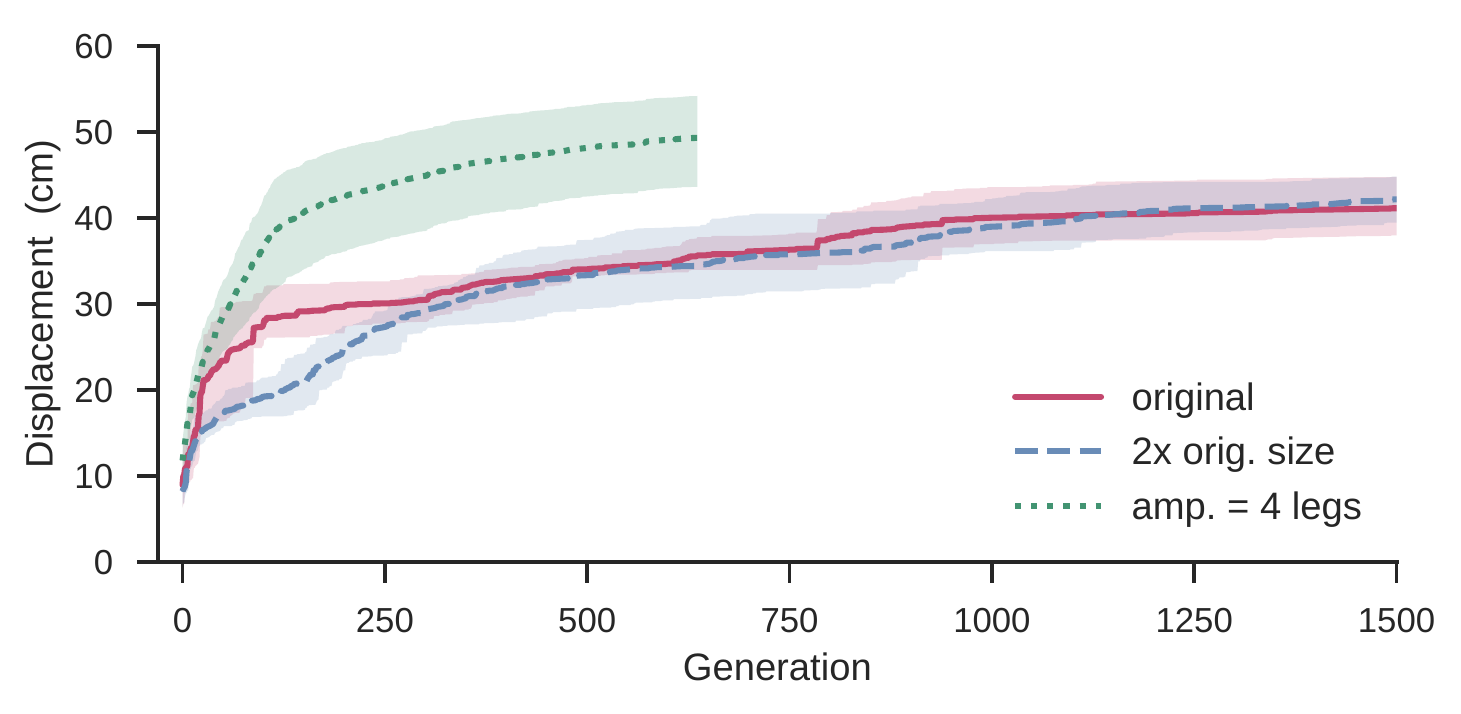}\\
\vspace{-6pt}
\caption{Mean displacement with 99\% CIs of controller optimization in the predamage quadruped isometrically enlarged to maximal volume, compared to shape optimization after amputation of all four legs.
}
\label{fig:size_effect}
\vspace{-2em}
\end{center}
\end{figure}

To explicitly control for the effect of body size, we optimized the controller of an otherwise identical quadruped that is twice the size of the original (Fig.~\ref{fig:size_effect}).
Isometrically increasing volume did not affect speed: There was no significant difference in speed (at the 0.01 level) between the enlarged and original quadruped. 
This is because the controller oscillations are added on top of (not relative to) the root shape~(Eq.~\ref{eq:beam_configuration}).
The enlarged robot has eight times the volume of the original, beam length oscillations still have the same amplitude.

Despite the fact that control was optimized for the original quadruped, and that amputation of all four legs removes 23\% of the original volume and actuation (Fig.~\ref{fig:scenarios}), robots that recovered from this particular insult through shape change (Fig.~\ref{fig:teaser}) move significantly faster than both the original and isometrically enlarged quadrupeds (Fig.~\ref{fig:size_effect}).
It follows that the efficacy of shapeshifting is not due simply to increased volume; rather, it is due to where and how the remnant structure's shape is deformed, which affects (e.g.) the robot's posture and mass distribution, its points of contact with the ground, and the storage and release of elastic strain energy, during locomotion.

\subsection{The found techniques of recovery.}

\begin{figure}[h!]
\begin{center}
\includegraphics[trim={0 0 0 0},clip,width=\linewidth]{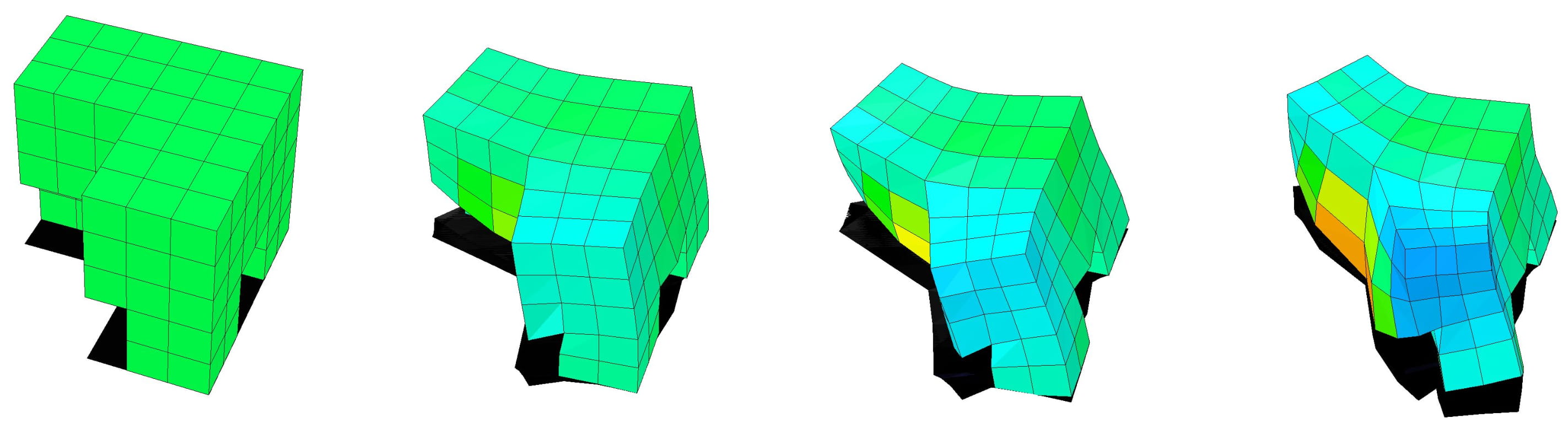}\\
\vspace{-4pt}
\caption{\label{fig:splay}
This damaged robot (amp.~=~1/4~body) contracted its hips and expanded its pelvis to recover function
(\href{https://youtu.be/UBvsR6tZf5c}{\textcolor{blue}{\textbf{\texttt{youtu.be/UBvsR6tZf5c}}}}).
}
\vspace{-1em}
\end{center}
\end{figure}

\begin{figure}[h!]
\begin{center}
\includegraphics[trim={0 0 0 0},clip,width=\linewidth]{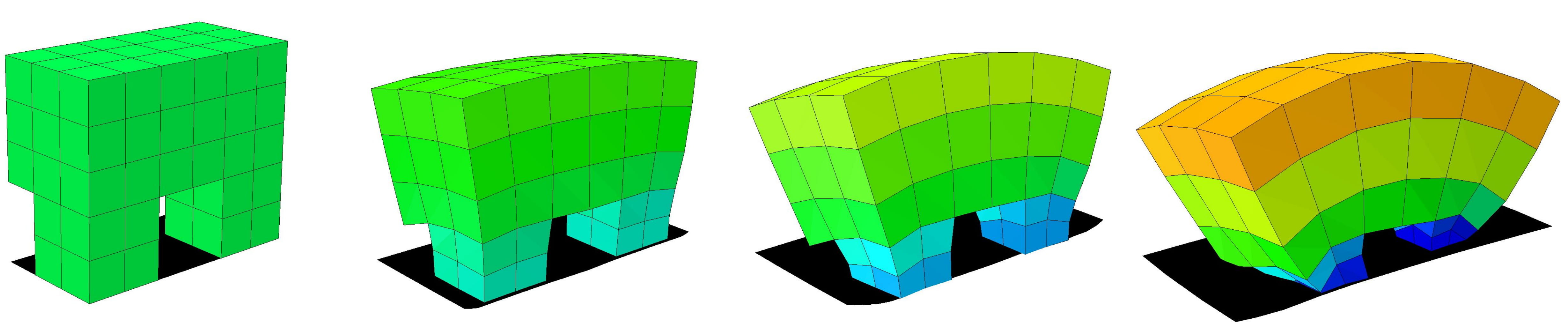}\\
\vspace{-4pt}
\caption{\label{fig:scruncher}
Shape change in this damage case (amp.~=~1/2~body) enabled upright, lengthwise movement, instead of falling over
(\href{https://youtu.be/nfCaVZVBmKI}{\textcolor{blue}{\textbf{\texttt{youtu.be/nfCaVZVBmKI}}}}).
}
\vspace{-1em}
\end{center}
\end{figure}

\begin{figure}[h!]
\begin{center}
\includegraphics[trim={0 0 0 0},clip,width=\linewidth]{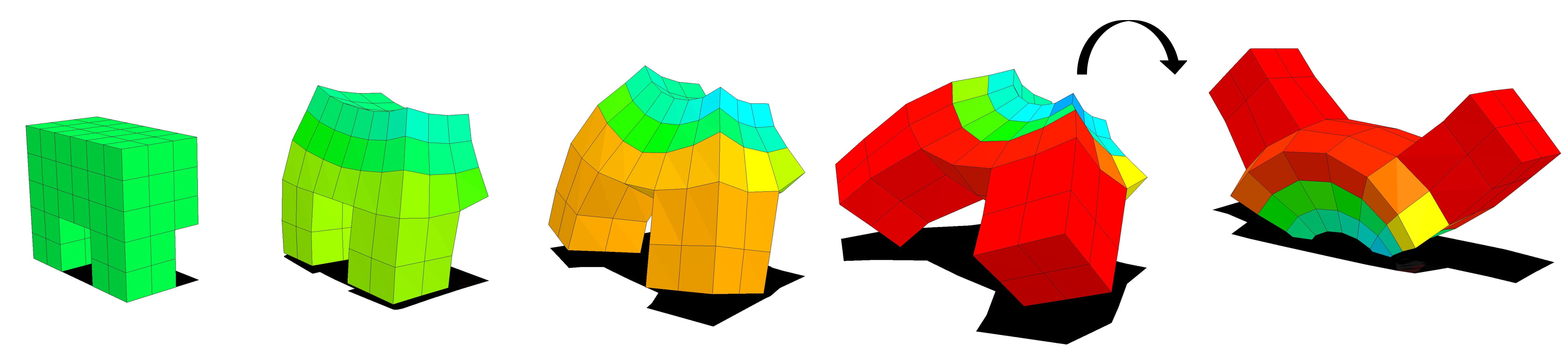}\\
\vspace{-5pt}
\caption{\label{fig:flipper}
This damaged robot (amp.~=~1/2~body) contracted its spine, expanded its limbs, and flipped over onto its back to walk lengthwise and exploit the elastic properties of its new arms
(\href{https://youtu.be/WwYdSnuJBBA}{\textcolor{blue}{\textbf{\texttt{youtu.be/WwYdSnuJBBA}}}}).
}
\vspace{-2em}
\end{center}
\end{figure}

We found that the optimizer discovered diverse recovery strategies through shape change (Figs.~\ref{fig:splay}-\ref{fig:flipper}), whereas controller readaptation often converged on the same strategy.
For example, with all four legs removed due to damage, the robot is reduced to a cuboid, and the only viable locomotion technique found by controller readaptation was crawling.
During postdamage optimization, many of these robots evolved crawling by peristalsis or in a manner that resembles the serpentine crawling of snakes \cite{alexander2003principles}.

Deforming the structure, even in a random manner, tends to produce greater frictional anisotropy which enhances peristalsis and serpentine crawling, and enables yet simpler forms of movement such as two-anchor crawling \cite{alexander2003principles}.

Nevertheless, crawling is inefficient because of drag.
Here, in many damage cases, behavioral competence was recovered through shape changes that partially or completely (but, due to material constraints, never perfectly) regenerated missing legs.
Notably, when all four legs were amputated, recovery strongly converged on the solution of regeneration, and the resulting designs were some of the fastest overall~(Fig.~\ref{fig:teaser}).
Note that the objective function does not assume or directly select for legged locomotion.

Other amputations, however, can be beneficial, if they result in a shape that is easier to efficiently control than the original:
Prior to damage, the robot's sagittal silhouette resembles a~$\Pi$.
When two adjacent legs are amputated, the resulting $\Gamma$ shape, which initially falls forward like this \rotatebox[origin=cB]{-60}{$\Gamma$} due to gravity, tends to rapidly surpass predamage performance through controller readaptation alone, despite its diminished size.

When only half of a leg was lost to injury, some robots contracted all of the undamaged legs to recover a stable but shorter quadrupedal form.
Others seemed to simply regenerate the missing part through local volumetric expansion at the site of damage:
The stump was isometrically expanded into a leg that was the same length as the original but much wider.

On closer inspection, however, many of those who regenerated a limb also made various other compensatory shape changes away from the site of damage, such as expanding and curving their spine.
Thus even when damage is isolated to a small part of the robot's structure,
global changes, in addition to local repairs, can sometimes streamline recovery.

When damage was distributed across a wider portion of the body, a diversity of solutions were discovered.
For example, after the amputation of a quarter of the robot's body, the robot occasionally splayed out its pelvis to form a straighter and faster shape~(Fig.~\ref{fig:splay}).
And, after the amputation of three legs, some robots once again grew replacements, but, because of other changes (e.g.,~a greatly expanded back),
the remaining ``genuine'' leg needed to be partially contracted and tucked inward for balance~(Fig.~\ref{fig:tucker}).

In the case where half of the robot's body is removed, the undeformed structure falls under gravity onto its side; one local optima was thus to crawl ``facedown''.
A better strategy was found in which the robot could remain upright by using the two remaining limbs as forelegs and expanding the stump to form a wide hind leg.
An equally proficient strategy was observed in which the robot diminished one or both of its legs, expanded its spine, and moved longitudinally~(Fig.~\ref{fig:scruncher}).

There were many successful variations on this theme,
but one of the best designs in this case did the exact opposite: The robot expanded its remaining limbs to their maximum volume, contracted its spine, and flipped over (once) to walk longitudinally with the added momentum generated from large, swaying front and back limbs~(Fig.~\ref{fig:flipper}).

However, after the most extreme insult, when all but a quarter of the robot is lost, there is insufficient material to regenerate legs or execute other more extreme shape changes. 
Neither recovery option cultivated (visually) appreciable gains in fitness.
Yet, while this case removes 71\% of the original volume it is significantly \textit{less} deleterious to controller optimization than amputating the four legs, which removes just 23\% of the original volume.
Insult is thus a matter of kind, not degree.

\subsection{The transferal of recovery strategies to reality.}

\begin{figure*}
\begin{center}
\includegraphics[trim={-40pt 0 0 0},clip,width=\linewidth]{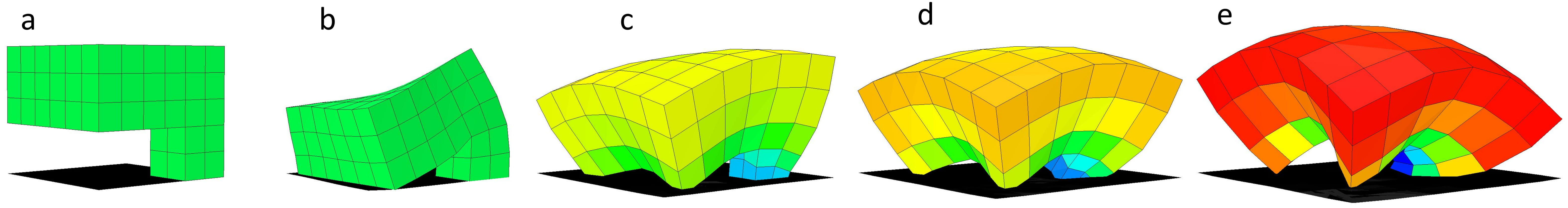} \\
\includegraphics[trim={0 0 0 -5pt},clip,width=\linewidth]{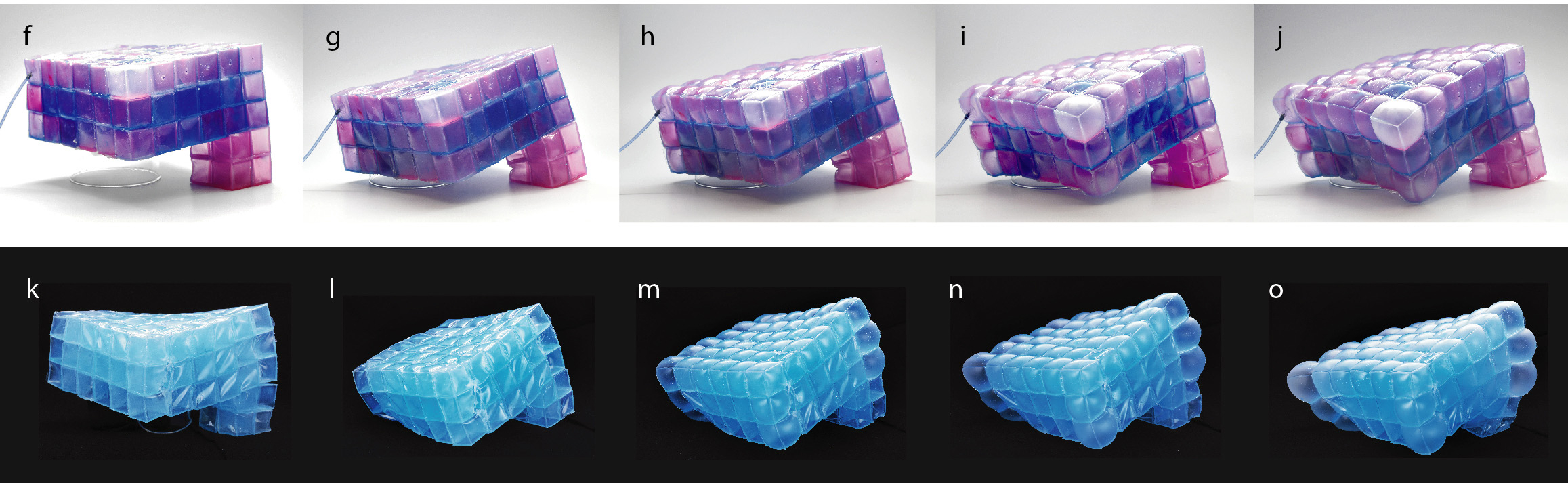}\\
\vspace{-2pt}
\caption{\label{fig:tucker}After losing three legs to injury (amp.~=~3~legs), the former quadruped is reduced to a monopedal structure~(\textbf{a}), the shape of which was then optimized for locomotion speed, resulting in an expanded spine, the folding-inward of the remnant predamage leg, and the ``regeneration'' of the three missing legs~(\textbf{c-e}).
This simulated strategy was then realized in two implementations using pneumatically-actuated, cubic elastomer bladders. The purple robot~(\textbf{f-j}) consists of two layers of drip-molded silicone; the blue robot~(\textbf{k-o}) consists of a single layer, and is thus less stable but more deformable.
A single air inlet here yields the rudiments of appropriate shape change, but pressure oscillations in this setup did not yield locomotion
(\href{https://youtu.be/A2KTGhCFxK8}{\textcolor{blue}{\textbf{\texttt{youtu.be/A2KTGhCFxK8}}}}).
}
\vspace{-2em}
\end{center}
\end{figure*}

To investigate the potential for directly transferring recovery strategies from simulation to reality, we aimed to transfer the overall shapes that are pictured in Figs.~\ref{fig:teaser} and~\ref{fig:tucker}. 
In these particular cases, the optimizer found shapes with contiguous sections of voxels actuated to similar levels. 
Thus, we here examine the one-actuator case, in which the voxels with the largest rest volumes|the top layer of ($6\times6=36$) voxels and the two corner voxels just below the top layer, on each corner of the torso (8 in total)|were connected to the same air inlet. 
Voxels not hooked into the air line were punctured to allow for passive deflation and contraction of the robot, mimicking the simulated robot's ability to contract voxels by decreasing the rest lengths~$b_i$. 


The purple robot adequately expands the top layer of voxels in both cases (Figs.~\ref{fig:teaser}j and~\ref{fig:tucker}j), but fails to reach the overall target shapes drawn in Figs.~\ref{fig:teaser}e and~\ref{fig:tucker}e. 
Although further increasing pressure did indeed lead to larger deformations,
the outer voxels inflate farther than interior ones, limiting the maximum viable actuation pressure. 
The thinner voxel walls of the blue robot exacerbated this issue~\mbox{(Fig.~\ref{fig:tucker}k-o)}, 
but their increased flexibility enabled a more faithful transferal of overall surface curvature.
Another limitation we discovered was friction.
Fully realizing the target shape in Fig.~\ref{fig:tucker}e requires the robot to drag its leg inward across the floor, tucking it under its body; but the silicone leg often stuck to the surface, preventing the prescribed maneuver in reality.

The silicone design and 1-axis rotational molding technique are still quite promising. 
Even when inflated at high enough pressures to make the outer voxels approximately spherical (Fig.~\ref{fig:tucker}o), the voxels did not rupture. 
To achieve more consistent expansion of interior and exterior voxels, the later should be inflated at a lower pressure than the former.
By incorporating strain sensors \cite{white_low-cost_2017} and closed-loop control in future, the robot could correct for this variation on the fly.
By actuating different voxels at different pressures, and enabling active contraction in addition to expansion, a much wider range of simulated shapes could be attained in reality.

\section{Discussion}
\label{sec:discussion}

In this paper, a new approach to robot damage recovery has been proposed.
Instead of presenting the remnant shape of the damaged robot to optimization as 
fixed,
we enable optimization to change this shape as the essential part of the recovery process.
In doing so we realized a machine that recovered more function than an otherwise equivalent system that could adapt its controller but not deform its shape.

In future work we will improve the transferal of simulated morphing machines
to physical ones using existing sim2real methods~\cite{bharadhwaj2018data, bongard2006resilient, cully2015robots, hwangbo2019learning, kwiatkowski2019task, tan2018sim} adapted appropriately to meet the additional
transferal demands dictated by soft materials~\cite{matas2018sim}. 
We will also generalize our optimization method
such that control and shape readaptation can be combined as dictated by the form of damage, predamage structure of the robot, and its task environment.

\subsection{Biological regeneration.}

In past work, rigid-bodied robots have been venerated for their ability to ``adapt like animals''~\cite{bongard2006resilient,cully2015robots}.
These machines, which were constructed from undeformable metals and hard plastics, 
automatically learned to control their bodies in spite of missing or broken legs.
But when an animal loses one or more of its legs to injury, it does not adapt by merely searching for a new mental representation of behavior that successfully maps onto the damaged body. 
Rather, they often fundamentally deform their damaged ``hardware'' into something more controllable.

Evidence for this abounds.
A famous example is the congenitally two-legged goat described by Slijper~\cite{slijper1942biologic}: 
an otherwise normal goat which was born without forelegs adopted an upright posture and learned to walk on its hind legs alone.
In addition to enlarged 
hind legs, 
striking changes in morphology were documented, including
a greatly elongated gluteal tongue and 
an innovative arrangement of small tendons,
a narrowed pelvis,
an oval (rather than V-shaped) thoracic cross-sectional shape,
a curved spine, 
and an unusually large neck~\cite{west2005developmental}.
The animal's body resembled that of a kangaroo more closely than that of a normal goat.

Other animals can regenerate.
The planarian flatworm can be cut into many pieces (the record is 279) all of which grow back to a full organism, regenerating not just tail and head, but eyes and the complete nervous system~\cite{montgomery1974minimal}.
Vertebrates, such as 
frogs, also display the capability of regenerating limbs, 
jaws, eyes and a variety of internal structures~\cite{brockes1997amphibian}. 
Humans too (especially children) are sometimes capable of fingertip regeneration after distal phalange amputation~\cite{illingworth1974trapped}. 

\subsection{Mechanisms of biological regeneration.}

Several of the mechanisms by which organisms achieve these forms of 
self-editing of their own anatomy pose design
challenges and future research directions for robotics.

First is the ability to harness the behavior of low-level components (cells) towards a specific large-scale goal-state: salamanders can regenerate whole limbs, eyes, tails, ovaries, and other organs \cite{mccusker2011axolotl},
but growth and remodeling ceases when a correctly shaped and sized organ is complete \cite{pezzulo2016top}.
Second is the flexibility and robustness of systems under novel conditions. For example, tadpoles whose facial organs are experimentally placed in abnormal configurations will undergo novel rearrangements to still give rise to normal frog faces during metamorphosis \cite{vandenberg2012normalized}, 
showing that the genome encodes not a hardwired set of movements for each organ but rather specifies a machine that can remodel toward the same target morphology from a variety of unexpected starting states. 
Thus, it is critical to understand and exploit the ability of evolution to give rise to hardware that is well-adapted to the normal environment but also retains significant plasticity \cite{sullivan2016physiological}. 

Third is the fact that during regeneration, the tissues making growth and morphogenesis decisions are themselves being drastically rearranged: thus, the computational control circuitry \textit{is itself} the object of the deformation actuators, forming a closed loop in which information is reliably processed in a medium that is constantly changing \cite{pezzulo2015re}. 
Finally, the remarkable robustness of morphological computation extends to information learned within the lifetime of the organism \cite{blackiston2015stability}.
Butterflies, which result from a caterpillar brain that is almost completely dissolved during metamorphosis, still remember information learned during the caterpillar stage \cite{blackiston2008retention}. 
Flatworms, which regrow their entire heads, still remember information they learned prior to decapitation \cite{corning1967regeneration, shomrat2013automated}. 

Attempts to implement these capabilities in artificial systems (whether robotic or via synthetic biology) are likely to enrich not only engineering technology, but also to feed back to the biological sciences and biomedicine. 
The current understanding of computation in biological tissues has numerous gaps, which are only likely to be filled by attempts to build these capabilities from the ground up \cite{kamm2018perspective}.

\subsection{Metamorphosing machines.}

It has been shown here that robots, too, are not only capable of regenerating limbs, but that such deformation can manifest by selecting for function recovery alone, instead of a target legged shape.

However, this ability largely depends on the material with which robots are made, for even if morphology is free to change in rigid bodies, the ways in which such change can occur are limited at best.
In~\cite{bongard2011morphological}, robots used a combination of rotary and linear actuators to slowly angle appendages downward and extrude them outward, thus simulating limb growth.
In softer machines, there are more ways for morphology to change: 
The soft robot used here was able to locally deform its geometry to bend, twist, compress or expand throughout its body.
Its also possible, although not investigated here, for soft robots to change their material properties, such as stiffness, 
through (e.g.) granular jamming~\cite{brown2010universal,kriegman2018interoceptive}. 

The possibility of this latter change highlights the inadequacy of the name ``soft robot''.
When a granular jamming robot jams (removes excess internal air to become stiff) does it cease to be a soft robot?
What if it never unjams?
For the purposes of damage repair, the most important property of soft robots is not that they are soft \textit{per se}, but that they may easily change their structural and material properties (possibly including stiffness).
One can envisage future ``rigid'' nanobots capable of self-assembling into a protean metamachine that can rearrange so as to regrow a lost part; but that day seems far off, whereas soft robots, capable of continuous morphological change, are already becoming a reality.

The future of this line of work promises not just new robotic systems but also new science. Shapeshifting robots, recast as scientific tools, can shed new light on old biological questions about developmental plasticity, regeneration and homeostasis~\cite{kriegman2017minimal,kriegman2018morphological,lobo2012modeling}.
And, symmetrically, new theories about the mechanisms that lie at the heart of such questions can be physically instantiated and optimized in a new breed of useful, autonomous and adaptive machines.

\section*{Acknowledgements}

This work was supported by NSF award EFRI-1830870 and
DARPA contract HR0011-18-2-0022.
Dylan Shah was supported by NASA 
STRF-80NSSC17K0164.
Computation was provided by the
Vermont Advanced Computing Core.
We thank Gabrielle Branin 
for the sim2real 
demonstration 
videos.

\bibliographystyle{plainnat}
\bibliography{main}

\begin{thebibliography}{46}
\providecommand{\natexlab}[1]{#1}
\providecommand{\url}[1]{\texttt{#1}}
\expandafter\ifx\csname urlstyle\endcsname\relax
  \providecommand{\doi}[1]{doi: #1}\else
  \providecommand{\doi}{doi: \begingroup \urlstyle{rm}\Url}\fi

\bibitem[Alexander(2003)]{alexander2003principles}
R~McNeill Alexander.
\newblock \emph{Principles of animal locomotion}.
\newblock Princeton University Press, 2003.
\newblock ISBN 9780691126340.

\bibitem[Bharadhwaj et~al.(2018)Bharadhwaj, Wang, Bengio, and
  Paull]{bharadhwaj2018data}
Homanga Bharadhwaj, Zihan Wang, Yoshua Bengio, and Liam Paull.
\newblock A data-efficient framework for training and sim-to-real transfer of
  navigation policies.
\newblock \emph{arXiv preprint arXiv:1810.04871}, 2018.
\newblock URL \url{https://arxiv.org/abs/1810.04871}.

\bibitem[Blackiston et~al.(2008)Blackiston, Casey, and
  Weiss]{blackiston2008retention}
Douglas~J Blackiston, Elena~Silva Casey, and Martha~R Weiss.
\newblock Retention of memory through metamorphosis: can a moth remember what
  it learned as a caterpillar?
\newblock \emph{PLoS One}, 3\penalty0 (3):\penalty0 e1736, 2008.
\newblock URL \url{https://doi.org/10.1371/journal.pone.0001736}.

\bibitem[Blackiston et~al.(2015)Blackiston, Shomrat, and
  Levin]{blackiston2015stability}
Douglas~J Blackiston, Tal Shomrat, and Michael Levin.
\newblock The stability of memories during brain remodeling: a perspective.
\newblock \emph{Communicative \& integrative biology}, 8\penalty0 (5):\penalty0
  e1073424, 2015.
\newblock URL \url{https://doi.org/10.1080/19420889.2015.1073424}.

\bibitem[Bongard(2011)]{bongard2011morphological}
Josh Bongard.
\newblock Morphological change in machines accelerates the evolution of robust
  behavior.
\newblock \emph{Proceedings of the National Academy of Sciences}, 108\penalty0
  (4):\penalty0 1234--1239, 2011.
\newblock URL \url{https://doi.org/10.1073/pnas.1015390108}.

\bibitem[Bongard et~al.(2006)Bongard, Zykov, and Lipson]{bongard2006resilient}
Josh Bongard, Victor Zykov, and Hod Lipson.
\newblock Resilient machines through continuous self-modeling.
\newblock \emph{Science}, 314\penalty0 (5802):\penalty0 1118--1121, 2006.
\newblock URL \url{https://doi.org/10.1126/science.1133687}.

\bibitem[Booth et~al.(2018)Booth, Case, White, Shah, and
  Kramer-Bottiglio]{booth_addressable_2018}
Joran~W Booth, Jennifer~C Case, Edward~L White, Dylan~S Shah, and Rebecca
  Kramer-Bottiglio.
\newblock An addressable pneumatic regulator for distributed control of soft
  robots.
\newblock In \emph{2018 IEEE International Conference on Soft Robotics
  (RoboSoft)}, pages 25--30. IEEE, 2018.
\newblock URL \url{https://doi.org/10.1109/ROBOSOFT.2018.8404892}.

\bibitem[Brockes(1997)]{brockes1997amphibian}
Jeremy~P Brockes.
\newblock Amphibian limb regeneration: rebuilding a complex structure.
\newblock \emph{Science}, 276\penalty0 (5309):\penalty0 81--87, 1997.
\newblock URL \url{https://doi.org/10.1126/science.276.5309.81}.

\bibitem[Brown et~al.(2010)Brown, Rodenberg, Amend, Mozeika, Steltz, Zakin,
  Lipson, and Jaeger]{brown2010universal}
Eric Brown, Nicholas Rodenberg, John Amend, Annan Mozeika, Erik Steltz,
  Mitchell~R Zakin, Hod Lipson, and Heinrich~M Jaeger.
\newblock Universal robotic gripper based on the jamming of granular material.
\newblock \emph{Proceedings of the National Academy of Sciences}, 107\penalty0
  (44):\penalty0 18809--18814, 2010.
\newblock URL \url{https://doi.org/10.1073/pnas.1003250107}.

\bibitem[Chatzilygeroudis et~al.(2018)Chatzilygeroudis, Vassiliades, and
  Mouret]{chatzilygeroudis2018reset}
Konstantinos Chatzilygeroudis, Vassilis Vassiliades, and Jean-Baptiste Mouret.
\newblock Reset-free trial-and-error learning for robot damage recovery.
\newblock \emph{Robotics and Autonomous Systems}, 100:\penalty0 236--250, 2018.
\newblock URL \url{https://doi.org/10.1016/j.robot.2017.11.010}.

\bibitem[Cheney et~al.(2013)Cheney, MacCurdy, Clune, and
  Lipson]{cheney2013unshackling}
Nick Cheney, Robert MacCurdy, Jeff Clune, and Hod Lipson.
\newblock Unshackling evolution: evolving soft robots with multiple materials
  and a powerful generative encoding.
\newblock In \emph{Proceedings of the Genetic and Evolutionary Computation
  Conference}, pages 167--174. ACM, 2013.
\newblock URL \url{https://doi.org/10.1145/2463372.2463404}.

\bibitem[Corning(1967)]{corning1967regeneration}
William~C Corning.
\newblock Regeneration and retention of acquired information.
\newblock In \emph{Chemistry of Learning}, pages 281--294. Springer, 1967.
\newblock URL \url{https://doi.org/10.1007/978-1-4899-6565-3_18}.

\bibitem[Cully et~al.(2015)Cully, Clune, Tarapore, and Mouret]{cully2015robots}
Antoine Cully, Jeff Clune, Danesh Tarapore, and Jean-Baptiste Mouret.
\newblock Robots that can adapt like animals.
\newblock \emph{Nature}, 521:\penalty0 503--507, 2015.
\newblock URL \url{https://doi.org/10.1038/nature14422}.

\bibitem[Eggenberger(1997)]{eggenberger1997evolving}
Peter Eggenberger.
\newblock Evolving morphologies of simulated 3{D} organisms based on
  differential gene expression.
\newblock In \emph{Proceedings of the European Conference on Artificial Life},
  pages 205--213, 1997.
\newblock URL \url{https://books.google.com/books?id=ccp8fzlyorAC}.

\bibitem[Goulding(2009)]{goulding2009circuits}
Martyn Goulding.
\newblock Circuits controlling vertebrate locomotion: moving in a new
  direction.
\newblock \emph{Nature Reviews Neuroscience}, 10\penalty0 (7):\penalty0 507,
  2009.
\newblock URL \url{https://doi.org/10.1038/nrn2608}.

\bibitem[Hiller and Lipson(2014)]{hiller2014dynamic}
Jonathan Hiller and Hod Lipson.
\newblock Dynamic simulation of soft multimaterial 3{D}-printed objects.
\newblock \emph{Soft Robotics}, 1\penalty0 (1):\penalty0 88--101, 2014.
\newblock URL \url{https://doi.org/10.1089/soro.2013.0010}.

\bibitem[Hwangbo et~al.(2019)Hwangbo, Lee, Dosovitskiy, Bellicoso, Tsounis,
  Koltun, and Hutter]{hwangbo2019learning}
Jemin Hwangbo, Joonho Lee, Alexey Dosovitskiy, Dario Bellicoso, Vassilios
  Tsounis, Vladlen Koltun, and Marco Hutter.
\newblock Learning agile and dynamic motor skills for legged robots.
\newblock \emph{Science Robotics}, 4\penalty0 (26), 2019.
\newblock URL \url{https://doi.org/10.1126/scirobotics.aau5872}.

\bibitem[Illingworth(1974)]{illingworth1974trapped}
Cynthia~M Illingworth.
\newblock Trapped fingers and amputated finger tips in children.
\newblock \emph{Journal of pediatric surgery}, 9\penalty0 (6):\penalty0
  853--858, 1974.
\newblock URL \url{https://doi.org/10.1016/S0022-3468(74)80220-4}.

\bibitem[Kamm et~al.(2018)]{kamm2018perspective}
Roger~D. Kamm et~al.
\newblock Perspective: The promise of multi-cellular engineered living systems.
\newblock \emph{APL Bioengineering}, 2\penalty0 (4):\penalty0 040901, 2018.
\newblock URL \url{https://doi.org/10.1063/1.5038337}.

\bibitem[Kano et~al.(2017)Kano, Sato, Ono, Aonuma, Matsuzaka, and
  Ishiguro]{kano2017brittle}
Takeshi Kano, Eiki Sato, Tatsuya Ono, Hitoshi Aonuma, Yoshiya Matsuzaka, and
  Akio Ishiguro.
\newblock A brittle star-like robot capable of immediately adapting to
  unexpected physical damage.
\newblock \emph{Royal Society open science}, 4\penalty0 (12):\penalty0 171200,
  2017.
\newblock URL \url{https://doi.org/10.1098/rsos.171200}.

\bibitem[Kriegman et~al.(2017)Kriegman, Cheney, Corucci, and
  Bongard]{kriegman2017minimal}
Sam Kriegman, Nick Cheney, Francesco Corucci, and Josh~C. Bongard.
\newblock A minimal developmental model can increase evolvability in soft
  robots.
\newblock In \emph{Proceedings of the Genetic and Evolutionary Computation
  Conference}, pages 131--138. ACM, 2017.
\newblock \doi{10.1145/3071178.3071296}.
\newblock URL \url{https://arxiv.org/abs/1706.07296}.

\bibitem[Kriegman et~al.(2018{\natexlab{a}})Kriegman, Cheney, and
  Bongard]{kriegman2018morphological}
Sam Kriegman, Nick Cheney, and Josh Bongard.
\newblock How morphological development can guide evolution.
\newblock \emph{Scientific reports}, 8\penalty0 (1):\penalty0 13934,
  2018{\natexlab{a}}.
\newblock URL \url{https://doi.org/10.1038/s41598-018-31868-7}.

\bibitem[Kriegman et~al.(2018{\natexlab{b}})Kriegman, Cheney, Corucci, and
  Bongard]{kriegman2018interoceptive}
Sam Kriegman, Nick Cheney, Francesco Corucci, and Josh~C. Bongard.
\newblock Interoceptive robustness through environment-mediated morphological
  development.
\newblock In \emph{Proceedings of the Genetic and Evolutionary Computation
  Conference}, pages 109--116. ACM, 2018{\natexlab{b}}.
\newblock \doi{10.1145/3205455.3205529}.
\newblock URL \url{https://arxiv.org/abs/1804.02257}.

\bibitem[Kwiatkowski and Lipson(2019)]{kwiatkowski2019task}
Robert Kwiatkowski and Hod Lipson.
\newblock Task-agnostic self-modeling machines.
\newblock \emph{Science Robotics}, 4\penalty0 (26), 2019.
\newblock URL \url{http://doi.org/10.1126/scirobotics.aau9354}.

\bibitem[Lobo et~al.(2012)Lobo, Beane, and Levin]{lobo2012modeling}
Daniel Lobo, Wendy~S Beane, and Michael Levin.
\newblock Modeling planarian regeneration: a primer for reverse-engineering the
  worm.
\newblock \emph{PLoS computational biology}, 8\penalty0 (4):\penalty0 e1002481,
  2012.
\newblock URL \url{https://doi.org/10.1371/journal.pcbi.1002481}.

\bibitem[Mahdavi and Bentley(2003)]{mahdavi2003evolutionary}
Siavash~Haroun Mahdavi and Peter~J Bentley.
\newblock An evolutionary approach to damage recovery of robot motion with
  muscles.
\newblock In \emph{European Conference on Artificial Life}, pages 248--255.
  Springer, 2003.
\newblock URL \url{https://doi.org/10.1007/978-3-540-39432-7_27}.

\bibitem[Matas et~al.(2018)Matas, James, and Davison]{matas2018sim}
Jan Matas, Stephen James, and Andrew~J Davison.
\newblock Sim-to-real reinforcement learning for deformable object
  manipulation.
\newblock In \emph{Proceedings of The 2nd Conference on Robot Learning},
  volume~87 of \emph{Proceedings of Machine Learning Research}, pages 734--743.
  PMLR, 2018.
\newblock URL \url{http://proceedings.mlr.press/v87/matas18a.html}.

\bibitem[McCusker and Gardiner(2011)]{mccusker2011axolotl}
Catherine McCusker and David~M Gardiner.
\newblock The axolotl model for regeneration and aging research: a mini-review.
\newblock \emph{Gerontology}, 57\penalty0 (6):\penalty0 565--571, 2011.
\newblock URL \url{https://doi.org/10.1159/000323761}.

\bibitem[Miller(2004)]{miller2004evolving}
Julian~Francis Miller.
\newblock Evolving a self-repairing, self-regulating, french flag organism.
\newblock In \emph{Proceedings of the Genetic and Evolutionary Computation
  Conference}, pages 129--139. Springer, 2004.
\newblock URL \url{https://doi.org/10.1007/978-3-540-24854-5_12}.

\bibitem[Montgomery and Coward(1974)]{montgomery1974minimal}
JR~Montgomery and SJ~Coward.
\newblock On the minimal size of a planarian capable of regeneration.
\newblock \emph{Transactions of the American Microscopical Society},
  93\penalty0 (3):\penalty0 386, 1974.
\newblock URL \url{https://doi.org/10.2307/3225439}.

\bibitem[Morin et~al.(2014)Morin, Kwok, Lessing, Ting, Shepherd, Stokes, and
  Whitesides]{morin_elastomeric_2014}
Stephen~A. Morin, Sen~Wai Kwok, Joshua Lessing, Jason Ting, Robert~F. Shepherd,
  Adam~A. Stokes, and George~M. Whitesides.
\newblock Elastomeric tiles for the fabrication of inflatable structures.
\newblock \emph{Advanced Functional Materials}, 24\penalty0 (35):\penalty0
  5541--5549, 2014.
\newblock URL \url{https://doi.org/10.1002/adfm.201401339}.

\bibitem[Pezzulo and Levin(2015)]{pezzulo2015re}
Giovanni Pezzulo and Michael Levin.
\newblock Re-membering the body: applications of computational neuroscience to
  the top-down control of regeneration of limbs and other complex organs.
\newblock \emph{Integrative Biology}, 7\penalty0 (12):\penalty0 1487--1517,
  2015.
\newblock URL \url{https://doi.org/10.1039/c5ib00221d}.

\bibitem[Pezzulo and Levin(2016)]{pezzulo2016top}
Giovanni Pezzulo and Michael Levin.
\newblock Top-down models in biology: explanation and control of complex living
  systems above the molecular level.
\newblock \emph{Journal of The Royal Society Interface}, 13\penalty0
  (124):\penalty0 20160555, 2016.
\newblock URL \url{http://doi.org/10.1098/rsif.2016.0555}.

\bibitem[Ren et~al.(2015)Ren, Chen, Dasgupta, Kolodziejski, W{\"o}rg{\"o}tter,
  and Manoonpong]{ren2015multiple}
Guanjiao Ren, Weihai Chen, Sakyasingha Dasgupta, Christoph Kolodziejski,
  Florentin W{\"o}rg{\"o}tter, and Poramate Manoonpong.
\newblock Multiple chaotic central pattern generators with learning for legged
  locomotion and malfunction compensation.
\newblock \emph{Information Sciences}, 294:\penalty0 666--682, 2015.
\newblock URL \url{https://doi.org/10.1016/j.ins.2014.05.001}.

\bibitem[Schmidt and Lipson(2011)]{schmidt2011age}
Michael Schmidt and Hod Lipson.
\newblock Age-fitness pareto optimization.
\newblock In \emph{Genetic Programming Theory and Practice VIII}, pages
  129--146. Springer, 2011.
\newblock URL \url{https://doi.org/10.1007/978-1-4419-7747-2_8}.

\bibitem[Shomrat and Levin(2013)]{shomrat2013automated}
Tal Shomrat and Michael Levin.
\newblock An automated training paradigm reveals long-term memory in planaria
  and its persistence through head regeneration.
\newblock \emph{Journal of Experimental Biology}, pages jeb--087809, 2013.
\newblock URL \url{http://doi.org/10.1242/jeb.087809}.

\bibitem[Slijper(1942)]{slijper1942biologic}
EJ~Slijper.
\newblock Biologic anatomical investigations on the bipedal gait and upright
  posture in mammals, with special reference to a little goat, born without
  forelegs.
\newblock In \emph{Proceedings of the Koninklijke Nederlandse Akademie Van
  Wetenschappen}, volume~45, pages 288--295, 407--415, 1942.

\bibitem[Stanley(2007)]{stanley2007compositional}
Kenneth~O Stanley.
\newblock Compositional pattern producing networks: A novel abstraction of
  development.
\newblock \emph{Genetic programming and evolvable machines}, 8\penalty0
  (2):\penalty0 131--162, 2007.
\newblock URL \url{https://doi.org/10.1007/s10710-007-9028-8}.

\bibitem[Sullivan et~al.(2016)Sullivan, Emmons-Bell, and
  Levin]{sullivan2016physiological}
Kelly~G Sullivan, Maya Emmons-Bell, and Michael Levin.
\newblock Physiological inputs regulate species-specific anatomy during
  embryogenesis and regeneration.
\newblock \emph{Communicative \& integrative biology}, 9\penalty0 (4):\penalty0
  e1192733, 2016.
\newblock URL \url{http://doi.org/10.1080/19420889.2016.1192733}.

\bibitem[Tan et~al.(2018)Tan, Zhang, Coumans, Iscen, Bai, Hafner, Bohez, and
  Vanhoucke]{tan2018sim}
Jie Tan, Tingnan Zhang, Erwin Coumans, Atil Iscen, Yunfei Bai, Danijar Hafner,
  Steven Bohez, and Vincent Vanhoucke.
\newblock Sim-to-real: Learning agile locomotion for quadruped robots.
\newblock In \emph{Proceedings of Robotics: Science and Systems}, 2018.
\newblock URL \url{https://doi.org/10.15607/RSS.2018.XIV.010}.

\bibitem[Vandenberg et~al.(2012)Vandenberg, Adams, and
  Levin]{vandenberg2012normalized}
Laura~N Vandenberg, Dany~S Adams, and Michael Levin.
\newblock Normalized shape and location of perturbed craniofacial structures in
  the xenopus tadpole reveal an innate ability to achieve correct morphology.
\newblock \emph{Developmental Dynamics}, 241\penalty0 (5):\penalty0 863--878,
  2012.
\newblock URL \url{http://doi.org/10.1002/dvdy.23770}.

\bibitem[West-Eberhard(2005)]{west2005developmental}
Mary~Jane West-Eberhard.
\newblock Developmental plasticity and the origin of species differences.
\newblock \emph{Proceedings of the National Academy of Sciences}, 102:\penalty0
  6543--6549, 2005.
\newblock URL \url{https://doi.org/10.1073/pnas.0501844102}.

\bibitem[White et~al.(2017)White, Yuen, Case, and Kramer]{white_low-cost_2017}
Edward~L. White, Michelle~C. Yuen, Jennifer~C. Case, and Rebecca~K. Kramer.
\newblock Low-cost, facile, and scalable manufacturing of capacitive sensors
  for soft systems.
\newblock \emph{Advanced Materials Technologies}, 2\penalty0 (9):\penalty0
  1700072, 2017.
\newblock URL \url{https://doi.org/10.1002/admt.201700072}.

\bibitem[White et~al.(2005)White, Zykov, Bongard, and Lipson]{white2005three}
Paul White, Victor Zykov, Josh Bongard, and Hod Lipson.
\newblock Three dimensional stochastic reconfiguration of modular robots.
\newblock In \emph{Proceedings of Robotics: Science and Systems}, 2005.
\newblock URL \url{http://doi.org/10.15607/RSS.2005.I.022}.

\bibitem[Zhao et~al.(2015)Zhao, Li, Elsamadisi, and
  Shepherd]{zhao_scalable_2015}
Huichan Zhao, Yan Li, Ahmed Elsamadisi, and Robert Shepherd.
\newblock Scalable manufacturing of high force wearable soft actuators.
\newblock \emph{Extreme Mechanics Letters}, 3:\penalty0 89--104, 2015.
\newblock URL \url{https://doi.org/10.1016/j.eml.2015.02.006}.

\bibitem[Zykov et~al.(2005)Zykov, Mytilinaios, Adams, and
  Lipson]{zykov2005robotics}
Victor Zykov, Efstathios Mytilinaios, Bryant Adams, and Hod Lipson.
\newblock Self-reproducing machines.
\newblock \emph{Nature}, 435\penalty0 (7039):\penalty0 163, 2005.
\newblock URL \url{https://doi.org/10.1038/435163a}.

\end{thebibliography}

\end{document}